\documentclass[10pt,twocolumn,letterpaper]{article}

\usepackage{wacv}
\usepackage{times}
\usepackage{epsfig}
\usepackage{graphicx}
\usepackage{amsmath}
\usepackage{amssymb}
\usepackage{multirow}
\usepackage{algorithm}
\usepackage{algpseudocode}
\usepackage{makecell}
\usepackage{enumitem}
\usepackage{pifont}
\usepackage{array}
\newcolumntype{P}[1]{>{\centering\arraybackslash}p{#1}}

\def\wacvPaperID{709} 

\wacvfinalcopy 

\ifwacvfinal
\def\assignedStartPage{9876} 
\fi

\ifwacvfinal
\usepackage[breaklinks=true,bookmarks=false]{hyperref}
\else
\usepackage[pagebackref=true,breaklinks=true,colorlinks,bookmarks=false]{hyperref}
\fi

\ifwacvfinal
\else
\fi

\begin{document}

\title{Multi-Modal Reasoning Graph for Scene-Text Based Fine-Grained Image Classification and Retrieval}


\author{Andres Mafla ~~ Sounak Dey ~~ Ali Furkan Biten ~~ Lluis Gomez ~~ Dimosthenis Karatzas\\
Computer Vision Center, UAB, Spain\\
{\tt\small \{andres.mafla, sdey, abiten, lgomez, dimos\}@cvc.uab.es}
}

\maketitle

\begin{abstract}
Scene text instances found in natural images carry explicit semantic information that can provide important cues to solve a wide array of computer vision problems.
In this paper, we focus on leveraging multi-modal content in the form of visual and textual cues to tackle the task of fine-grained image classification and retrieval. First, we obtain the text instances from images by employing a text reading system. Then, we combine textual features with salient image regions to exploit the complementary information carried by the two sources. Specifically, we employ a Graph Convolutional Network to perform multi-modal reasoning and obtain relationship-enhanced features by learning a common semantic space between salient objects and text found in an image. By obtaining an enhanced set of visual and textual features, the proposed model greatly outperforms previous state-of-the-art in two different tasks, fine-grained classification and image retrieval in the Con-Text\cite{karaoglu2013text} and Drink Bottle\cite{bai2018integrating} datasets.
\end{abstract}

\section{Introduction}

Since the advent of written text to represent ideas, 
humans have employed it to communicate non-trivial and semantically rich information. Nowadays, text can be found in an ubiquitous manner in images and video, especially in urban and man-made environments\cite{veit2016coco,karatzas2015icdar}. 
Extracting and analyzing such textual information in images jointly with the visual content, is indispensable to achieve full scene understanding.
In this work, we explore the role of such multi-modal cues, specifically in the form of visual and textual features to solve the task of fine-grained image classification and retrieval. 
\begin{figure}[htb]

\centering
\includegraphics[width=0.9\columnwidth]{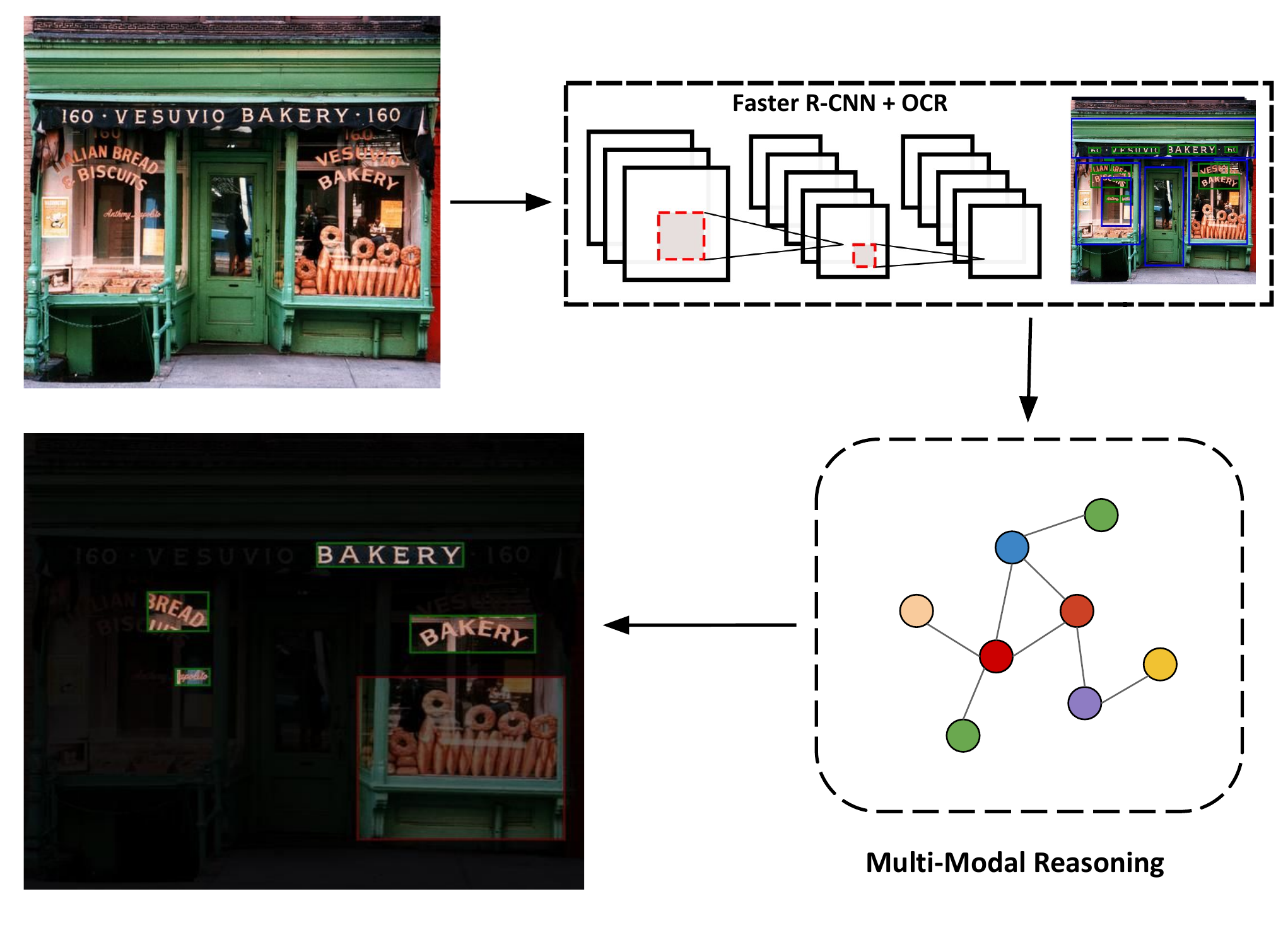}
\caption{The proposed model employs a Graph-based Multi-Modal Reasoning (MMR) module to enrich location-based visual and textual features in a combined semantic representation. The network learns at the output of the MMR to map strong complementary regions of visual (red) and text (green) instances to obtain discriminative features to perform fine-grained image classification and retrieval.}

\label{fig:gcn_overall}
\end{figure}

The task of fine-grained image classification (FGIC) consists of labeling a set of images that are visually alike. A lot of research on this problem has been oriented to differentiate visually similar objects such as birds\cite{ge2016temporal}, aircrafts \cite{maji2013fine}, and dog breeds\cite{khosla2011novel} among others, which more often than not require domain specific knowledge.
However, differentiating objects by leveraging available textual instances in the scene is an omnipresent practice in daily life. In this work, we focus on exploiting scene-text as the main discriminatory feature to perform FGIC.
A seminal work on leveraging textual cues was presented by Movshovitz \etal~\cite{movshovitz2015ontological}, who showcased that in order to classify store-fronts, a trained Convolutional Neural Network (CNN) had automatically learned to focus on scene text instances as the sole way to solve the given task. In the case of blurred or occluded text instances, the classification task is extremely challenging for humans as well. 
In this manner, scene text found in an image serves as an additional discriminative signal that a model should incorporate into its design. 
Additional research devoted to leveraging textual cues in the task of FGIC has been explored. Similar to our work, Karaoglu \etal ~\cite{karaoglu2013text,karaoglu2017words} introduces a simple pipeline to perform fine-grained classification using scene text and extending the previous work, 
an attention mechanism is proposed by Bai \etal~\cite{bai2018integrating} to learn a common semantic space. In a different approach, Mafla \etal~\cite{mafla2020fine} learns a morphological space by using textual instances as discriminative features rather than semantics to solve this task.

Departing from previous approaches, we exploit a structural representation between the studied modalities. Our work summarized in Figure~\ref{fig:gcn_overall} with publicly available code at 
\footnote{\url{https://github.com/AndresPMD}}, focuses on learning an enhanced visual representation that incorporates reasoning between salient regions of an image and scene text to construct a semantic space over which fine-grained classification is performed. In this example, we can observe that relevant regions such as the text "Bakery" and "Bread" are associated with a visual region that depicts pastry, both important cues to classify the given image.
Additionally, we show experiments of fine-grained image retrieval, using the same multi-modal representation, in the two evaluated datasets. Overall, the main contributions can be summarized as following:
\begin{itemize}[noitemsep,nolistsep]
\item We propose a novel architecture that greatly surpasses previous state-of-the-art results in two datasets by more than $5$\% on fine-grained classification and $10$\% on image retrieval by considering text and visual features of an image.
\item We design a fully end-to-end trainable pipeline that incorporates a Multi-Modal Reasoning module that combines textual and visual features that
do not rely on ensemble models or pre-computed features. 
\item We provide exhaustive experiments in which we analyze the effectiveness of different modules in our model architecture and the importance of scene text towards comprehensive models of image understanding.
\end{itemize}


\section{Related Work}
\label{sec:Related_work}
\subsection{Scene Text Detection and Recognition}
Localizing and recognizing text instances found in a natural image is a challenging problem due to the variability, orientation, occlusion and background noise among other factors~\cite{chen2020text}.
Deep learning based methods began with the work proposed by~\cite{jaderberg2016reading} which focused on a sliding window and a CNN to filter the proposals. The proposals were used as input into another CNN that posed the task as a classification problem over a large fixed dictionary of words. Later works take object detection pipelines such as YOLO~\cite{redmon2016yolo9000} used by~\cite{gupta2016synthetic} to obtain a Fully Convolutional Neural Network along with a focus on generating synthetic training data, which later became the go-to data to train text detectors and recognizers. Along these lines, a variation of SSD~\cite{liu2016ssd} is presented by~\cite{liao2017textboxes, liao2018textboxes++}
to develop a text detector which easily integrates with a module trained for recognition. Methods that focus on an end-to-end recognition have been explored by~\cite{borisyuk2018rosetta} based on Faster R-CNN~\cite{ren2015faster}, which performs text detection and incorporates a Connectionist Temporal Classification (CTC)~\cite{graves2006connectionist} to recognize a given text instance. Similarly,~\cite{he2018end} presents a CNN as a region-based feature extractor, which are fed to two attention-based Long-Short Term Memories (LSTM) to predict bounding boxes and recognize the textual proposals. Multi-lingual models have been proposed as in the case of~\cite{buvsta2018e2e}, work that uses a CNN as an encoder and a CTC to decode the characters from a set of different languages. 

On a different approach, the Pyramidal Histogram of Characters (PHOC)~\cite{almazan2014word} is used to represent words and it has been amply used in text spotting in documents~\cite{sudholt2017learning} and in text retrieval in natural images~\cite{Gomez_2018_ECCV}. 
Despite all the progress done in scene text detection and recognition, it remains as an open problem in the computer vision community, with a special focus placed lately on multi-oriented text localization and recognition.

\subsection{Fine-Grained Classification}
The task of Fine-Grained Image Classification (FGIC) focuses on finding discriminative visual regions that often require domain specific knowledge to correctly perform the labeling task~\cite{wei2019deep}. Different to solely visual based FGIC methods, there has been growing interest to use textual cues to achieve this task by incorporating two modalities.

Closely related to this work, the initial approach taken by~\cite{karaoglu2017words} was to extract scene text and construct a bag of words, while the visual features where obtained by employing a pre-trained GoogLeNet~\cite{szegedy2015going}. Soon after,~\cite{bai2018integrating} proposes the usage of Textboxes~\cite{liao2017textboxes} to read scene-text in an image, a CNN to obtain visual features along with an attention mechanism  and a concatenation of the final features to learn a semantic space suitable for scene-text based FGIC. Later work performed by~\cite{mafla2020fine} employs a CNN as a visual feature extractor and uses the PHOC representation of a word along with the Fisher Vector~\cite{perronnin2007fisher} to learn a space based on the morphology of text instances to overcome Optical Character Recognition (OCR) errors. 
Several fusion methods are explored in the work by~\cite{mafla2020fine} but finally a concatenation of features is performed to solve the task of image classification and retrieval.

\begin{figure*}[ht]
\begin{center}
\includegraphics[width=0.95\linewidth]{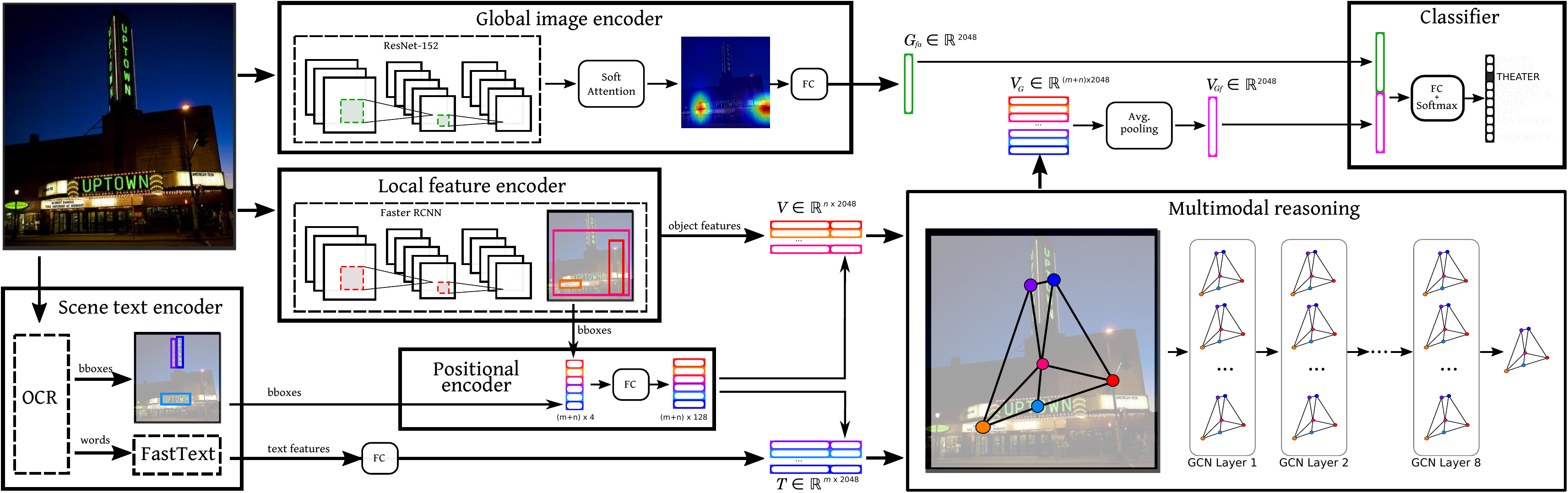}
\end{center}
\caption{Detailed model architecture. The proposed model combines features of regions of scene text and visual salient objects by employing a graph-based Multi-Modal Reasoning (MMR) module. The MMR module enhances semantic relations between the visual regions and uses the enriched nodes along with features from the Global Encoder to obtain a set of discriminatory signals for fine-grained classification and retrieval.}
\label{fig:model}
\end{figure*}

\subsection{Multi-Modal Fusion and Reasoning}
Several fusion-based techniques such as Multimodal Compact Bilinear Pooling (MCB)~\cite{gao2016compact, fukui2016multimodal}, Low-rank Bilinear Attention Network (MLB)~\cite{kim2016hadamard} and Block~\cite{ben2019block} have been explored to model relationships between language and vision. 
To model these interaction, attention-based~\cite{bahdanau2014neural} approaches also have been have been proposed~\cite{anderson2018bottom, xu2015show, kazemi2017show}. With the aim of designing models capable of reasoning, the intrinsic synergy between visual and textual features has been explored. Work such as~\cite{zellers2019recognition, li2019visual} employ variations of an LSTM and a Gated Recurrent Unit (GRU) to perform reasoning in a sequential manner. However, significant advances have been made by the usage of Graph Convolutional Networks (GCN)~\cite{kipf2016semi}, due to the proven capability of modelling relationships~\cite{schlichtkrull2018modeling} between nodes in a given graph. Along this road, GCNs have been successfully used in tasks that require reasoning such as VQA~\cite{narasimhan2018out, singh2019strings, gao2020multi}, image captioning~\cite{li2019know, yang2019auto} and image-sentence retrieval~\cite{li2019visual, liu2020graph}.

In this work, we propose a method to learn a richer set of visual features and model a more discriminative semantic space by employing a GCN. To the best of our knowledge, this is the first approach that integrates multimodal sources which come in the form of visual along textual features jointly with positional encoding into a GCN pipeline that performs reasoning for the task of scene-text based fine-grained image classification and retrieval. 
\section{Method}

In this section, we detail each of the components that comprise the proposed architecture. Figure~\ref{fig:model} depicts the overall scheme of the proposed model, which is formed by $6$ different modules: global image encoder, local feature encoder, text encoder, positional encoder, multi-modal reasoning graph and classification module. The local feature encoder employs features extracted based on the regions of interest obtained by a Faster R-CNN~\cite{ren2015faster} in a similar manner as the bottom-up attention model~\cite{anderson2018bottom}. The scene text encoder uses an OCR model to obtain scene text and further embed it into a common space. The goal is to obtain multi-modal node representations that leverage the semantic relationships found between salient objects and text instances within an image that are discriminative enough to perform fine-grained classification.

\subsection{Global Image Encoder}
We employ a CNN as an encoder, which in our case is a ResNet-152~\cite{he2016deep} pre-trained on ImageNet~\cite{deng2009imagenet} to acquire global image features. Particularly, given an image $I$ we take the output features 
before the last average pooling layer, which output is denoted as 
$G_f = \psi(I)$. In order to obtain a more descriptive set of global features and due to its differentiable properties, we compute a soft attention mechanism on top of the global features. This self-attention mechanism yields an attention mask, $attn_{mask}$, that assigns weights on different regions of the input image. The attention weights are learned in an end-to-end manner by convolving $1\times1$ kernels projected into a single dimensional filter and later followed by a Softmax function. In order to obtain the final attended global features, the attention mask is broadcasted and multiplied with the global features, which result is added to the global features $G_f$ to later be used as input of a Fully-Connected layer, $FC$, in the form of:
\begin{equation}
G_{fa} = FC( G_f + (G_f \times attn_{mask}) )
\end{equation}
where $G_{fa}\in \mathbb{R}^{1\times D}, G_f \in \mathbb{R}^{HxWxD}, attn_{mask} \in \mathbb{R}^{HxW}$ stands for the final encoded global features, where $D = 2048$, $H=7$ and $W=7$.

\subsection{Local Feature Encoder}

Following~\cite{anderson2018bottom}, we employ a Faster R-CNN~\cite{ren2015faster} pre-trained on Visual Genome~\cite{krishna2017visual} as the extractor of local visual features.
This approach allows us to obtain salient image regions that are potentially discriminative for our task. We use an IOU threshold of $0.7$ and a confidence threshold of $0.3$, and sort the obtained predictions before the last average pooling layer to use the top $n$ most confident regions of interest. Thus, we can represent the output of an image $I$ with a set of region features $R_f = \{ (r_1, bbox_{r_1}) ..., (r_n, bbox_{r_n}) \}, r_i \in \mathbb{R}^d $, where $r_i$ is the $i^{th}$ region of interest and $bbox_{r_i}$ is the $r_i$'s corresponding bounding box coordinates normalized with respect to the image.
In our experiments, we set $n=36$ and  the obtained features have a dimension of $d=2048$.
In order to encode the local visual features, we project the features through a fully-connected layer.

In this manner we obtain the final encoded local features that will serve as input to the multi-modal GCN in the form of $V_f = \{v_1, ..., v_n \}, v_i \in \mathbb{R}^D $, where $D=1920$ is the dimension of the final embedding space. The bounding boxes obtained to represent these regions are later used as input into the positional encoder module. If there are less than $n = 36$ regions in an image, a zero padding scheme is adopted.

\subsection{Text Encoder}
To extract text contained in an image, we ran several public state of the art text recognizers as well as a commercial OCR model provided by Google\footnote{\url{https://cloud.google.com/vision/}}. We extract the transcriptions of each word, denoted as $w_i$, as well as the corresponding bounding boxes, $bbox_{w_i}$.
In particular, we extract the top $m$ most confident textual instances found in an image. The transcriptions are embedded using Fasttext~\cite{bojanowski2017enriching} and the bounding boxes will be used as input in the positional encoder branch. We employ the FastText embedding due to its capability of encoding word morphology in the form of n-grams as well as preserving a semantic space similar to Word2Vec~\cite{mikolov2013distributed} while at the same time dealing with out of vocabulary words. Analogously to the local features case, we project the obtained embedded textual features by passing them through a fully-connected layer. 
The final textual features are represented by $T_f = \{t_1, ..., t_m \}, t_i \in \mathbb{R}^D $, where $D=1920$ is the dimension of the final embedding space and $m = 15$ is the number of text proposals extracted from an image. In the case that there is no text found in a given image, similarly to the local encoder module, zero padding is employed.

\subsection{Positional Encoder}
Encoding the position of objects and text instances within an image can provide important relational information about the scene. For example text found on top of a building often refers to its class in a explicit manner contrary to text found in any other location in the image. To meet this end, we design a positional encoding that takes as input a predicted bounding box of an object or text instance. The input to the positional encoder describes the top left $(x_1, y_1)$, and bottom right $(x_2, y_2)$ coordinates normalized according to the image size, and is a concatenation of the bounding boxes of the local and text regions of interest. The bbox matrix is given by: $bboxes_{input} = \{bbox_{r_1},.., bbox_{r_n}, bbox_{t_1},..., bbox_{t_m}\}$ where $bbox_i = (x_1, y_1, x_2, y_2)$.
In order to encode them, we pass the bounding boxes over a fully-connected in a similar way as the same as previous sections. 
The final encoded representation can be described as:  $bboxes = \{bbox_{r_1},.., bbox_{r_n},bbox_{t_1},...,bbox_{t_m}\}, bbox_i \in \mathbb{R}^b$, in which the dimension $b = 128$ represents the final encoded bounding boxes.

\subsection{Multi-modal Reasoning Graph}
Due to the showcased capability of graphs to describe reasoning between objects~\cite{singh2019strings, yang2019auto, gao2020multi, liu2020graph}, we construct a richer set of region-based visual descriptors that exploit the semantic correlation between visual and textual features. In order to do so, we initialize the node features as local visual features and textual features concatenated with their respective positional encoding of bounding boxes.
We can describe the node features as:

\begin{align*}
V=\{ (v_1, bbox_{r_1}), ..(v_n, bbox_{r_n}), (t_1, bbox_{t_1}), ...\\
...,(t_m, bbox_{t_m})\}, V \in \mathbb{R}^{(n+m)\times D}
\end{align*}
where $n, m$ is the number of visual and textual features, respectively. In our case, $n+m = 51$ and  $D = 1920+128 = 2048$. Furthermore, we construct the affinity matrix $R$ which measures the degree of correlation of between two visual regions. The construction of the affinity matrix is given by:
\begin{equation}
R_{ij} =  \phi(k_i)^T  \gamma(k_j) 
\end{equation}
where $k_i, k_j \in V$, $\phi(.)$ and $\gamma(.)$ are two fully connected layers that are learned end-to-end by back propagation at training time.
If we define $k = n+m$, then the obtained affinity matrix consists of a shape $k\times k$.
Once $R$ is calculated, we can define our graph by $G = (V,R) $, in which the nodes are represented by the local and textual features $V$ and the edges are described by $R$. The obtained graph describes through the affinity matrix $R$ the degree of semantic and spatial correlation between two nodes. We use the formulation of Graph Convolutional Networks given by~\cite{kipf2016semi} to obtain a reasoning over the nodes and edges. Particularly, we use residual connections in the GCN formulation as it is presented by~\cite{li2019visual}. We can write the equation that describes a single Graph Convolution layer performed as:
\begin{equation}
V^{l}_g = W_r^{l}(R^{l} V^{l-1} W_g^{l}) + V^{l-1}
\end{equation}
where $R \in \mathbb{R}^{k \times k}$ is the affinity matrix , $V \in \mathbb{R}^{k \times D}$ the local visual features, $W_g  \in \mathbb{R}^{D \times D}$ is a learnable weights matrix of the GCN, $W_r\in \mathbb{R}^{k \times k}$ corresponds to the residual weights matrix and $l$ is the number of GCN layer.
Notice that passing $V$ through the GCN layer, a richer set of multi-modal features is obtained. In order to find an enhanced representation of the visual features we apply $l=8$ GCN layers in total, which finally yields a set of enriched nodes that represent the visual features $V_{G}$ such that:
\begin{align*}
V_G =\{ v_{g1}, ..,v_{gk} \}, V_G \in \mathbb{R}^{k \times D}
\end{align*}

\subsection{Classification}
In order to combine the global $G_{fa}$ and the enriched local and textual $V_G$ visual features, firstly we perform an average pooling of the $V_G$ tensor. Specifically, we can rewrite the final local feature vector $V_{Gf}$ as:
\begin{equation}
V_{Gf} = \frac{1}{k}\sum_{n=1}^{k} V_{gi}
\end{equation}

Lastly, we simply concatenate the two obtained vectors $V_{Gf}$ and $G_{fa}$, to obtain the final vector $F$ that is used as input for the final fully-connected layer for classification denoted by: $F = [G_{fa}, V_{Gf}]$

By applying a softmax to the output of the final layer, we obtain a probability distribution of a class label given an input image. The model is trained in an end-to-end fashion optimized with the cross entropy loss function described by:

\begin{equation}
\label{e:loss_function}
J(\theta) = -\frac{1}{N}\sum_{n=1}^{N}\sum_{i=1}^{\mathcal{C}} y_i^{n} log ({p}_i^{n})
\end{equation}

Where, $C$ is the number of classes, $N$ the dataset samples such that each pair contains an annotation $\{x^{(n)},y^{(n)} \} | n = 1,2,...,N $, and $p^n$ is the predicted output label.
 
\section{Experiments and Results}
This section presents an introduction to the datasets employed in this work, as well as the implementation details, ablation studies performed, and a thorough analysis of the results obtained in the experiments conducted.
\subsection{Datasets}

The Con-Text dataset was introduced by Karaoglu \etal~\cite{karaoglu2013text} and is a subset of ImageNet~\cite{deng2009imagenet}, constructed by selecting the sub-categories of "building" and "place of business". This dataset contains $24,255$ images in total divided into three-folds to divide training and testing sets. This dataset introduces $28$ visually similar categories of images such as Cafe, Pizzeria, and Pharmacy in which in order to perform fine-grained classification, text is a necessary cue to solve otherwise a very difficult task even for humans. 
This dataset closely resembles natural circumstances due to the fact that the images are taken without considering scene text instances, thus some images do not have text present in them.

The Drink Bottle dataset was presented by Bai \etal~\cite{bai2018integrating} and as the Con-Text dataset, it is a subset of images of ImageNet~\cite{deng2009imagenet}, specifically taken from the sub-categories of soft drink and alcoholic drink. The dataset is divided in three-folds as well and contains $18,488$ images. There are $20$ image categories which include visually similar instances such as Coca Cola, Pepsi Cola and Cream Soda. Akin to the Con-Text dataset, some images contain scene-text while others do not have it.

\subsection{Implementation Details}
In our experiments in order to extract visual regions of an image, we use the same settings as~\cite{anderson2018bottom}. We take the top $n=36$ ROIs and encode them along with their bounding boxes into a common space of $2048$-d. The transcribed text is sorted by confidence score and we take the top $m=15$ confident predictions. We embed the textual instances by a using a pre-trained FastText model with $1$ million $300_d$ word vectors, trained with sub-word information on Wikipedia2017, UMBC webbase corpus and statmt.org news dataset. The obtained $300$-d textual vectors are projected with the corresponding bounding boxes into a $2048$-d space. The Faster R-CNN~\cite{ren2015faster} from~\cite{anderson2018bottom} and the OCR models, both employed as initial feature extractor modules use pre-trained weights and are not updated at training stage. The rest of the weights of each module in the model are learned in an end-to-end manner during training.
The graph-based multimodal reasoning module employs $8$ multi-modal GCN layers to obtain the final enriched visual features. 
In the last full-connected layer before classification, we employ a dropout rate of 0.3 to avoid over-fitting on the evaluated datasets. In general, we employ Leaky ReLU as an activation function in all layers except the last one, in which we use a Softmax to compute the class label probabilities. The proposed model is trained for $45$ epochs, but an early stop condition is employed. We use a combination of optimizers comprised by RAdam~\cite{liu2019radam} and Lookahead~\cite{zhang2019lookahead}. The batch size employed in all our experiments is $64$, with a starting learning rate of $0.001$ that decays by a factor of $0.1$ on the epochs $15$, $30$ and $45$. The momentum value used on the optimizers is $0.9$ and the weight decay is $0.0005$. 

\subsection{Comparison with the State-of-the-Art}

We show the experimental results of our method compared to previous state-of-the-art on Table~\ref{tab:context_bottles_short}. We can note that the performance obtained in the Con-Text significantly surpasses the previous best performing method by $5.9$\%. The improvement in the Drink-Bottle dataset is more modest, of about $1.98$\%, 
however it is still significant.

\begin{table}[h]
\begin{center}
\small
\begin{tabular}{l|llcc}

\textbf{Method} & \textbf{OCR}   & \textbf{Emb.} & \textbf{Context} & \textbf{Bottles} \\ \hline
Karao.\cite{karaoglu2013text}        & Custom         & BoB$^1$    & $39.0$                    & $-$                     \\
Karao.\cite{karaoglu2017words}        & Jaderberg & Probs$^2$   & $77.3$                    & $-$                     \\
Bai\cite{bai2018integrating}            & Textboxes      & GloVe              & $78.9$                    & $-$                     \\
Bai\cite{bai2018integrating}$^\dagger$             & Textboxes      & GloVe              & $79.6$                    & $72.8$                  \\
Bai\cite{bai2018integrating}$^\dagger$             & Google OCR      & GloVe              & $80.5$                    & $74.5$                  \\
Mafla\cite{mafla2020fine}           & SSTR-PHOC      & FV      & $80.2$                    & $77.4$                  \\
Proposed            & E2E-MLT        & Fasttext           & $82.36$                   & $78.14$                 \\
Proposed            & SSTR-PHOC      & PHOC               & $82.77$                   & $78.27$                 \\
Proposed            & SSTR-PHOC      & FV      & $83.15$                   & $77.86$                 \\\hline
\textbf{Final}            & \textbf{Google OCR}     & \textbf{Fasttext}           & \boldmath$85.81$                   & \boldmath$79.87$                
\end{tabular}
\end{center}
\caption{Classification performance of state-of-the art methods on the Con-Text and Bottles dataset. The results depicted with $^\dagger$ are based on an ensemble model. The embeddings labeled as: $^1$ refer to a Bag of Bigrams, $^2$ is a probability vector along a dictionary. The acronym FV stands for Fisher Vector. The metric depicted is the mean Average Precision (mAP in \%).}
\label{tab:context_bottles_short}
\end{table}

We believe the improvement is greater in Con-Text due to the text instances found in it, which refer mostly to business places without particular out of vocabulary words, therefore a semantic space for classification is more discriminative when compared to the Drink-Bottle dataset.
To provide further insights, we conducted experiments by employing the final model along with different OCRs and word embeddings in both datasets. It is essential to note that state-of-the-art results are achieved by the usage of other OCRs as well, showing that the proposed pipeline still outperforms previous methods. 
Results showing the classification scores of each evaluated class and further analysis are shown in the Supplementary Material section.

When comparing to previous methods, it is worth revisiting previous approaches. The results reported by~\cite{bai2018integrating} used an ensemble of classifiers to reach the obtained performance. As an additional experiment to showcase the effect of using the same OCR as our proposed model is included, and it shows that our model vastly outperforms the evaluated pipeline not because of the OCR system employed. On the other side, the work done by~\cite{mafla2020fine} requires offline pre-computation of the Fisher Vector by training a Gaussian Mixture Model and tuning the hyper-parameters involved. In this manner, the method proposed in this work do not require an ensemble and the features used are learned in an end-to-end manner at training time. We clearly show that the proposed pipeline surpasses other approaches even when employing a set of different scene-text OCRs. 

With the aim of offering additional insights, we present in Table~\ref{tab:text_visual_performance} the performance of previous state of the art methods compared with our proposed method in a subset of the test set such that the evaluated images either contain scene-text or not. The results show the average performance along the $3$ different splits of each dataset. We can observe that our model is able to perform better than previous approaches in both scenarios while a more significant improvement is achieved in images that contain scene-text, which we treat as the major discriminative feature to perform the task of fine-grained classification.

\begin{table}[t!]
\begin{center}
\small

\begin{tabular}{c|rr|rr}
\multirow{2}{*}{\textbf{Method}} & \multicolumn{2}{c|}{\textbf{Context}}                                   & \multicolumn{2}{c}{\textbf{Bottles}}                                  \\
                                 & \multicolumn{1}{l}{\textbf{I + T}} & \multicolumn{1}{l|}{\textbf{I - T}} & \multicolumn{1}{l}{\textbf{I + T}} & \multicolumn{1}{l}{\textbf{I - T}} \\ \hline
Bai~\cite{bai2018integrating}                              & $78.92$                               & $71.63$                               & $71.61$                              & $62.25$                             \\
Mafla~\cite{mafla2020fine}                           & $77.94$                             & $72.59$                              & $78.57$                              & $68.97$                             \\ \hline
\textbf{Ours}                    & \boldmath$86.76$                     & \boldmath$74.31$                     & \boldmath$82.75$                     & \boldmath$69.19$                   
\end{tabular}

\end{center}
\caption{Classification performance of the proposed method on the subset of images from the test set of the Con-Text and Bottles dataset such that the images: contain scene-text (I + T) and do not contain scene-text (I - T) . The metric depicted is the mean Average Precision (mAP in \%).}
\label{tab:text_visual_performance}
\end{table}


\subsection{Importance of Textual Features}

In order to assess the importance of the scene text found in images, we follow the previous works~\cite{karaoglu2017words, bai2018integrating, mafla2020fine} by defining two different evaluation baselines, the visual features based and the textual features based. Moreover, due to the fact that the evaluated datasets do not contain text transcriptions as ground truth, we evaluated the effectiveness of the OCR employed in the fine-grained classification task.

The visual only evaluates all the test set images by only employing the global encoder features $G_f$ in the first scenario and the global encoder along with the self attention features $G_{fa}$ in the second scenario. In both cases the output of the global encoder, a $2048$-d feature vector, is directly passed through a fully connected layer to obtain the final classification prediction. In the textual only, the baselines are evaluated only in the subset of images which contained spotted scene text. The results of each baseline by employing visual only, different OCRs and word embeddings are shown in Table~\ref{tab:results1}.
\begin{table}[]
\begin{center}
\small
\begin{tabular}{c|l|c|c}
\multicolumn{1}{l|}{}       & \textbf{Model}         & \textbf{Context} & \textbf{Bottles} \\ \hline
\multirow{2}{*}{\textbf{Visual}} & CNN            & $62.11$             & $65.15$            \\
                            & CNN + Self Attention            & $63.78$              & $66.62$            \\ \hline
\multirow{7}{*}{\textbf{Textual}} 
& Texspotter+w2v$^\dagger$        & $35.09$             & $50.68$            \\
& Texspotter+glove$^\dagger$       & $34.52$             & $50.26$            \\
& Texspotter+fasttext$^\dagger$    & $36.71$              & $51.93$             \\
& E2E\_MLT+w2v$^\dagger$           & $44.36$             & $43.98$            \\
& E2E\_MLT+glove$^\dagger$         & $44.25$             & $42.64$            \\
& E2E\_MLT+fasttext$^\dagger$      & $45.07$             & $44.31$            \\
 & FOTS+w2v    & $43.22$              & $41.33$             \\
 & FOTS+glove    & $43.71$              & $41.85$             \\
 & FOTS+fasttext    & $44.19$              & $42.69$             \\
 & Google OCR+w2v    & $53.87$              & $53.47$             \\
 & Google OCR+glove    & $54.48$              & $54.39$             \\
 & \textbf{Google OCR+fasttext}    & \boldmath$55.61$              & \boldmath$55.16$             \\
& PHOC$^\dagger$          & $49.18$    & $52.39$   \\ 
\cline{2-4}
\multicolumn{1}{l|}{}       & \textbf{Fisher Vector (PHOC)$^\dagger$} & \boldmath$63.93$    & \boldmath$62.41$ \\ 
\hline
\end{tabular}
\end{center}
\caption{Visual only and Textual only results. The textual only results were performed on the subset of images that contained spotted text. The results with $^\dagger$ were reported by~\cite{mafla2020fine}. The metric depicted is the mean Average Precision (mAP in \%).}
\label{tab:results1}
\end{table}

Following a previous approach~\cite{mafla2020fine}, we employ $m=15$ text instances and pre-trained word embeddings that yield $300$-d vectors in the case of Word2Vec~\cite{mikolov2013distributed}, GloVe~\cite{pennington2014glove} and FastText~\cite{bojanowski2017enriching}. The textual tensor obtained is used as input to a fully connected layer, which output is used for classification purposes. In our experiments we evaluate two additional state-of-the-art scene text recognizers, FOTS~\cite{liu2018fots} and the commercially used Google OCR Cloud Vision based on an API. We note that the embedding that performs the best is Fasttext due to the capability of embedding out of vocabulary words by using character n-grams. 
Regarding the results, it was found that the best performing standard recognizer is the Google OCR, which employs a more compact ($300$-d) vector compared to a PHOC or a Fisher Vector. The PHOC embedding employs a $604$-d feature vector along with $m=15$ and the Fisher Vector is a single $38400$-d vector in our experiments. Overall, by using only textual features, the Fisher Vector based on PHOCs remains as the best performing descriptor. However, besides the high dimensional vector employed, extensive offline pre-computation is required to obtain such descriptor. Nonetheless, as it can be seen in Table~\ref{tab:context_bottles_short}, the FV descriptor does not achieve the best results in our final model.

\subsection{Ablation studies}

In this section, we present the incremental improvements and the effects obtained by the addition of each module that comprises the final architecture in the method proposed. 
\newline
Table~\ref{tab:quantitative_res} shows the quantitative results of adding components in the baseline model. Namely, we evaluate the effect of using self-attention and the multi-modal reasoning (MMR) module. We successively add to the attended global features ($G_{fa}$), local features ($V_f$), textual features ($T_f$) and the bounding boxes ($bboxes$) of both used in the Positional Encoder. 
\begin{table}[]
\begin{center}
\small
\begin{tabular}{lcc}
\multicolumn{1}{l|}{\textbf{Features}}       & \multicolumn{1}{c|}{\textbf{Context}} & \textbf{Bottles}                \\ \hline
\multicolumn{1}{l|}{$G_f$}                   & \multicolumn{1}{c|}{$62.11$}             & $65.15$                           \\ 
\multicolumn{1}{l|}{$G_{fa}$}                & \multicolumn{1}{c|}{$63.78$}              & $66.62$                           \\ \hline
\multicolumn{3}{c}{\textbf{without MMR}}                                               \\ \hline
\multicolumn{1}{l|}{$G_{fa}$ + $V_f$}            & \multicolumn{1}{c|}{$70.48$}             & $73.21$                           \\
\multicolumn{1}{l|}{$G_{fa}$ + $V_f$ + $T_f$}        & \multicolumn{1}{c|}{$78.72$}             & $76.43$                           \\
\multicolumn{1}{l|}{$G_{fa}$ + $V_f$ + $T_f$ + $bboxes$} & \multicolumn{1}{c|}{$80.12$}                  & $77.51$                                \\ \hline
\multicolumn{3}{c}{\textbf{with MMR}}                                                                      \\ \hline

\multicolumn{1}{l|}{$G_{fa}$ + $V_f$}            & \multicolumn{1}{c|}{$72.88$}             & $74.96$                           \\
\multicolumn{1}{l|}{$G_{fa}$ + $V_f$ + $T_f$}        & \multicolumn{1}{c|}{$82.51$}             & $77.46$                           \\
\multicolumn{1}{l|}{$V_f$ + $T_f$ + $bboxes$}                & \multicolumn{1}{c|}{$84.33$}                  & $75.42$                                 \\
\multicolumn{1}{l|}{\boldmath{$G_{fa}$ + $V_f$ + $T_f$ + $bboxes$}} & \multicolumn{1}{c|}{\boldmath$85.81$}             & \boldmath$79.87$
\end{tabular}
\end{center}
\caption{ Quantitative results of the different components that form the proposed model. $G_f$: Global features, $G_{fa}$: $G_f$ + Self-Attention, $V_f$: Local Features, $T_f$: Text Features, $bboxes$: Bounding Box information used by the Positional Encoder, MMR: Multi-modal Reasoning. Results are shown in terms of the mAP(\%).}
\label{tab:quantitative_res}
\end{table}
In order to assess the effectiveness of the multi-modal reasoning graph module, we compare a model that uses the Faster R-CNN ROIs without the usage of the MMR. It is observed that solely by using the Faster R-CNN features, an important boost is achieved. One of the biggest improvements is reached by the usage of scene text, which enforces the idea that textual information is essential to successfully discriminate between visually similar classes. By the incorporation of scene text, an improvement of $9.7$\% is gained in Con-Text and $2.5$\% in the Drink-Bottle datasets. Nonetheless, the improvement is accentuated by the usage of the MMR module, which produces as output richer local and textual features coming from the graph nodes. Finally by adding the positional encoder module into the MMR, another increase in the results is achieved. 
This encourages us to think that the MMR module learns relationships coming from semantic and spatial information. Insights into the attention masks learned and the reasoning coming from the MMR by using visual and textual regions can be found in the Supplementary Material section.
\setlength\tabcolsep{1.2pt}

\begin{figure*}[!htp]
\begin{center}
\small
\begin{tabular}{p{0.12\linewidth} p{0.12\linewidth} p{0.12\linewidth} p{0.12\linewidth}p{0.12\linewidth} p{0.12\linewidth} p{0.12\linewidth} p{0.12\linewidth}}
    \includegraphics[width=\linewidth,height=0.75\linewidth]{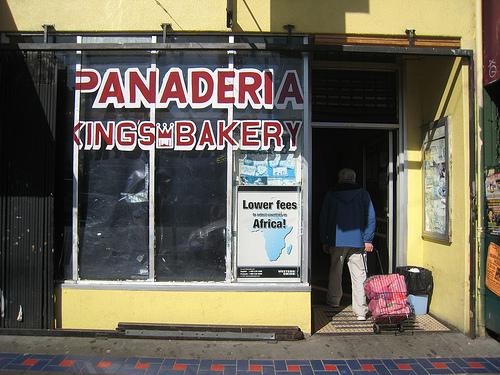}
    & 
    \includegraphics[width=\linewidth,height=0.75\linewidth]{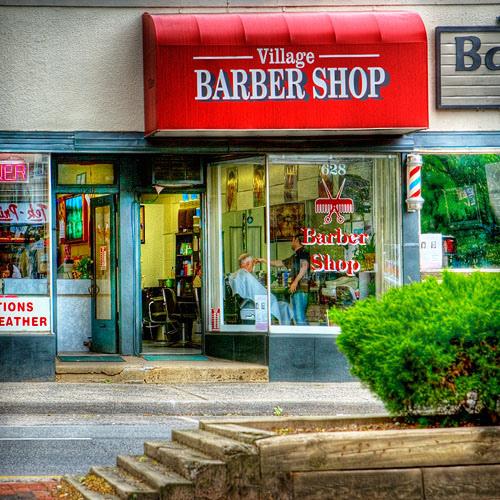}
    & 
    \includegraphics[width=\linewidth,height=0.75\linewidth]{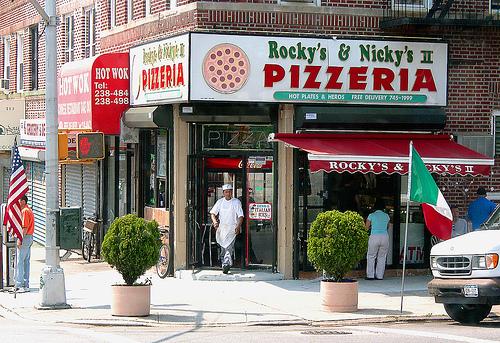}
    & 
    \includegraphics[width=\linewidth,height=0.75\linewidth]{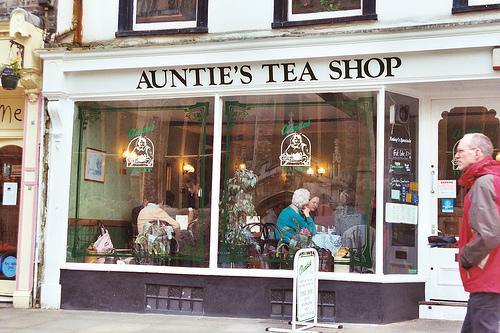}
    &
    \includegraphics[width=\linewidth,height=0.75\linewidth]{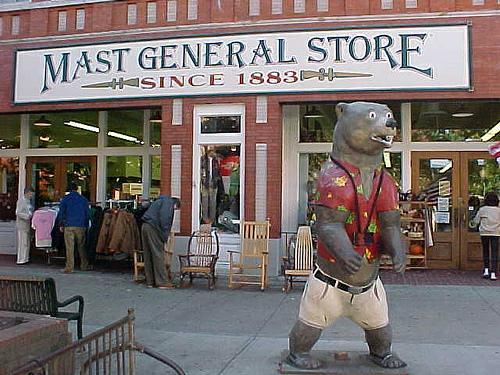} 
    & 
    \includegraphics[width=\linewidth,height=0.75\linewidth]{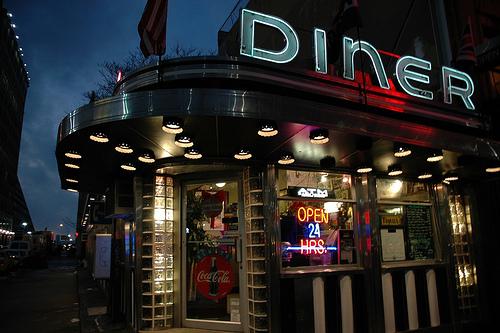}
    & 
    \includegraphics[width=\linewidth,height=0.75\linewidth]{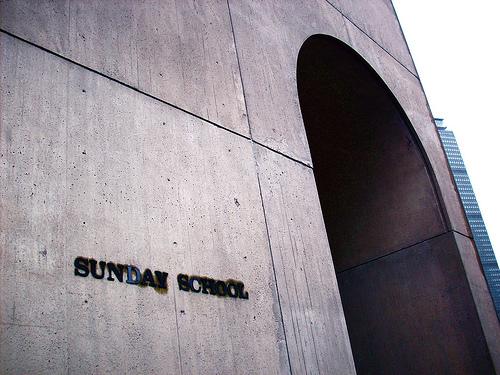}
    & 
    \includegraphics[width=\linewidth,height=0.75\linewidth]{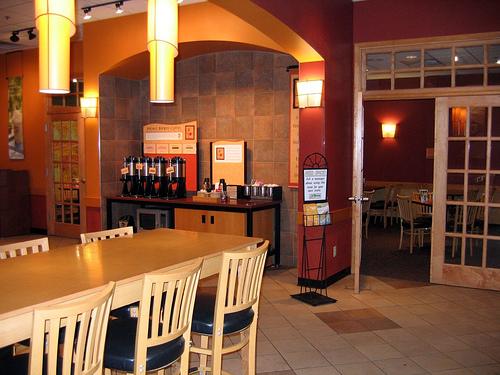}
    \\
     \footnotesize{\fontfamily{qhv}\selectfont \textbf{GT: } Bakery} \par {\color{blue}\footnotesize{\fontfamily{qhv}\selectfont \textbf{Bakery: 0.44}}}
    \par {\footnotesize{\fontfamily{qhv}\selectfont {Barber: 0.32}}}
    \par {\footnotesize{\fontfamily{qhv}\selectfont {Cafe: 0.15}}}
    &
     \footnotesize{\fontfamily{qhv}\selectfont \textbf{GT: } Barber} \par {\color{blue}\footnotesize{\fontfamily{qhv}\selectfont \textbf{Barber: 0.99}}}
    \par {\footnotesize{\fontfamily{qhv}\selectfont {Packing: 3.4e-7}}}
    \par {\footnotesize{\fontfamily{qhv}\selectfont {Discount: 1.8e-7}}}
    &
     \footnotesize{\fontfamily{qhv}\selectfont \textbf{GT: } Pizzeria} \par {\color{blue}\footnotesize{\fontfamily{qhv}\selectfont \textbf{Pizz.: 0.99}}}
    \par {\footnotesize{\fontfamily{qhv}\selectfont {Restaur: 4.9e-5}}}
    \par {\footnotesize{\fontfamily{qhv}\selectfont {Dinner: 1.2e-5}}}
    &
     \footnotesize{\fontfamily{qhv}\selectfont \textbf{GT: } Tea House} \par {\color{blue}\footnotesize{\fontfamily{qhv}\selectfont \textbf{Tea H: 0.98}}}
    \par {\footnotesize{\fontfamily{qhv}\selectfont {Cafe: 1.4e-2}}}
    \par {\footnotesize{\fontfamily{qhv}\selectfont {Barber: 1.3e-3}}}
    &
    \footnotesize{\fontfamily{qhv}\selectfont \textbf{GT: } Country S.} \par {\color{blue}\footnotesize{\fontfamily{qhv}\selectfont \textbf{CountryS: 0.94}}}
    \par {\footnotesize{\fontfamily{qhv}\selectfont {Tea H: 1.6e-2}}}
    \par {\footnotesize{\fontfamily{qhv}\selectfont {Cafe: 1.4e-2}}}
    &
     \footnotesize{\fontfamily{qhv}\selectfont \textbf{GT: } Diner} \par {\color{blue}\footnotesize{\fontfamily{qhv}\selectfont \textbf{Diner: 0.99}}}
    \par {\footnotesize{\fontfamily{qhv}\selectfont {Packing: 1.4e-7}}}
    \par {\footnotesize{\fontfamily{qhv}\selectfont {Restaur: 9.8e-8}}}
    &
     \footnotesize{\fontfamily{qhv}\selectfont \textbf{GT: } School} \par {\color{red}\footnotesize{\fontfamily{qhv}\selectfont \textbf{Theatre: 0.22}}}
    \par {\footnotesize{\fontfamily{qhv}\selectfont {Pharma: 0.18}}}
    \par {\footnotesize{\fontfamily{qhv}\selectfont {Barber: 0.18}}}
    &
     \footnotesize{\fontfamily{qhv}\selectfont \textbf{GT: } Cafe} \par {\color{red}\footnotesize{\fontfamily{qhv}\selectfont \textbf{Restaur: 0.79}}}
    \par {\footnotesize{\fontfamily{qhv}\selectfont {Packing: 0.11}}}
    \par {\footnotesize{\fontfamily{qhv}\selectfont {Bistro: 3.8e-2}}}

    \\
    & \\

    \includegraphics[width=\linewidth,height=0.75\linewidth]{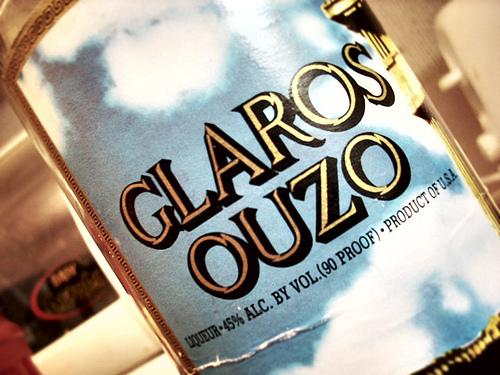} &
    \includegraphics[width=\linewidth,height=0.75\linewidth]{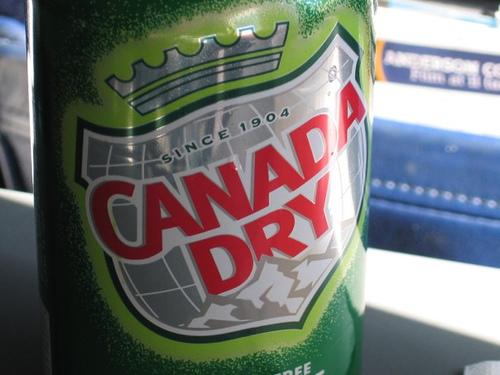} &
    \includegraphics[width=\linewidth,height=0.75\linewidth]{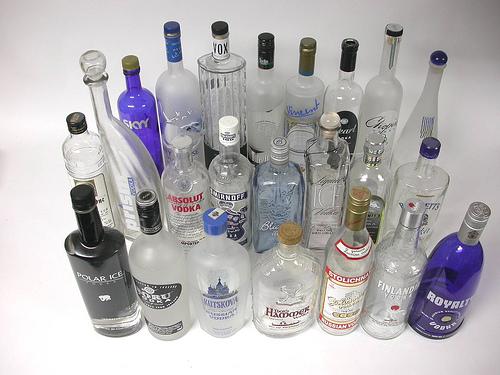} &
    \includegraphics[width=\linewidth,height=0.75\linewidth]{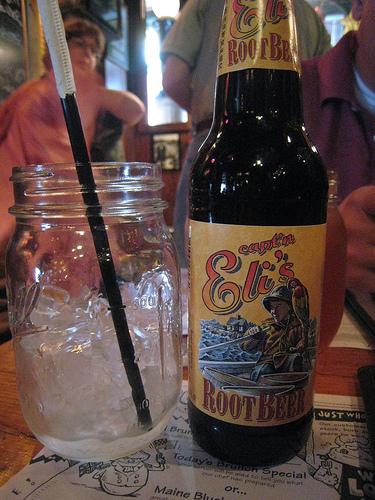} &
    \includegraphics[width=\linewidth,height=0.75\linewidth]{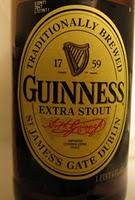} &
    \includegraphics[width=\linewidth,height=0.75\linewidth]{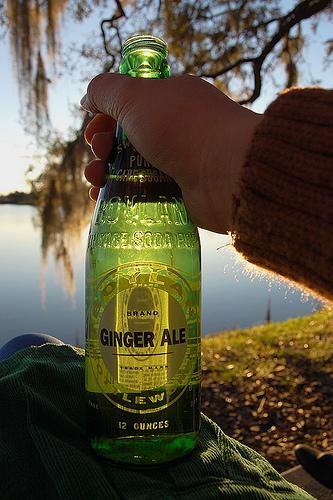} &
    \includegraphics[width=\linewidth,height=0.75\linewidth]{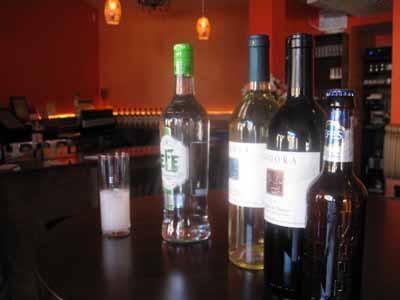} &
    \includegraphics[width=\linewidth,height=0.75\linewidth]{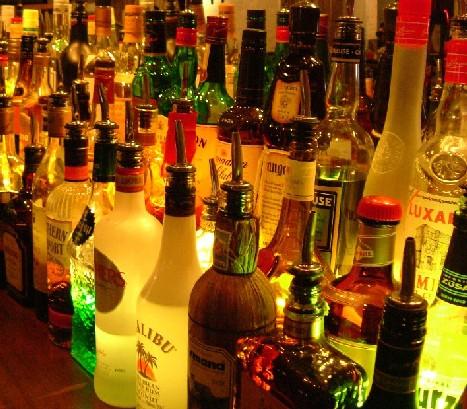}\\
    
    \footnotesize{\fontfamily{qhv}\selectfont \textbf{GT: } Ouzo} \par      {\color{blue}\footnotesize{\fontfamily{qhv}\selectfont \textbf{Ouzo: 0.99}}}
    \par {\footnotesize{\fontfamily{qhv}\selectfont {Bitter: 7e-5}}}
    \par {\footnotesize{\fontfamily{qhv}\selectfont {RootB: 1.5e-5}}}
    &
    \footnotesize{\fontfamily{qhv}\selectfont \textbf{GT: } GingerA} \par {\color{blue}\footnotesize{\fontfamily{qhv}\selectfont \textbf{GingerA: 0.99}}}
    \par {\footnotesize{\fontfamily{qhv}\selectfont {QuinW: 4.7e-3}}}
    \par {\footnotesize{\fontfamily{qhv}\selectfont {Sarsap: 6.2e-4}}}
    &
    \footnotesize{\fontfamily{qhv}\selectfont \textbf{GT: } Vodka} \par {\color{blue}\footnotesize{\fontfamily{qhv}\selectfont \textbf{Vodka: 0.99}}}
    \par {\footnotesize{\fontfamily{qhv}\selectfont {Ouzo: 2.9e-4}}}
    \par {\footnotesize{\fontfamily{qhv}\selectfont {QuinW: 1.3e-6}}}
    &
     \footnotesize{\fontfamily{qhv}\selectfont \textbf{GT: } RootB} \par {\color{blue}\footnotesize{\fontfamily{qhv}\selectfont \textbf{RootB: 0.99}}}
    \par {\footnotesize{\fontfamily{qhv}\selectfont {GingerA: 1.4e-3}}}
    \par {\footnotesize{\fontfamily{qhv}\selectfont {BirchB: 6.2e-4}}}
    &
    \footnotesize{\fontfamily{qhv}\selectfont \textbf{GT: } Guiness} \par      {\color{blue}\footnotesize{\fontfamily{qhv}\selectfont \textbf{Guiness: 0.99}}}
    \par {\footnotesize{\fontfamily{qhv}\selectfont {GingerA: 2.1e-6}}}
    \par {\footnotesize{\fontfamily{qhv}\selectfont {Ouzo: 1.2e-6}}}
    &
    \footnotesize{\fontfamily{qhv}\selectfont \textbf{GT: } GingerA} \par {\color{blue}\footnotesize{\fontfamily{qhv}\selectfont \textbf{GingerA: 0.99}}}
    \par {\footnotesize{\fontfamily{qhv}\selectfont {Ouzo: 1.8e-5}}}
    \par {\footnotesize{\fontfamily{qhv}\selectfont {CreamS: 5.9e-6}}}
    &
    \footnotesize{\fontfamily{qhv}\selectfont \textbf{GT: } Ouzo} \par {\color{red}\footnotesize{\fontfamily{qhv}\selectfont \textbf{Drambuie: 0.53}}}
    \par {\footnotesize{\fontfamily{qhv}\selectfont {Ouzo: 0.31}}}
    \par {\footnotesize{\fontfamily{qhv}\selectfont {Vodka: 8.8e-2}}}
    &
     \footnotesize{\fontfamily{qhv}\selectfont \textbf{GT: } Drambuie} \par {\color{red}\footnotesize{\fontfamily{qhv}\selectfont \textbf{Chablis: 0.29}}}
    \par {\footnotesize{\fontfamily{qhv}\selectfont {Vodka: 0.25}}}
    \par {\footnotesize{\fontfamily{qhv}\selectfont {Bitter: 0.13}}}

\end{tabular}
\end{center}
\caption{Classification predictions. The top-3 probabilities of a class are shown as well as the Ground Truth label performed on the test set. Without recognizing textual instances some images are extremely hard to classify even for humans. Text in blue and red is used to show correct and incorrect predictions respectively. Best viewed in color.}
\label{fig:long}
\vspace{-0.3cm}
\end{figure*}

Furthermore, we explore in our work several projection and fusion methods which are shown in Table~\ref{tab:projection_fusion}. In our experiments, Projection refers to the strategy used to reduce the dimensionality of the output tensor coming from the MMR as ${V_G}$ to obtain a single vector $V_{Gf}$. Late Fusion showcases the method employed to combine the features coming from $V_{Gf}$ and $G_{fa}$. Due to several works showing performance gains by the usage of attention~\cite{you2016image, xu2015show} and Recurrent Neural Networks~\cite{li2019visual, chen2019figure} as reasoning modules, we explored those alternatives, however no improvements were found. In the same manner, as it is presented by~\cite{mafla2020fine}, we explored two additional fusion mechanisms, MLB~\cite{kim2016hadamard} and Block~\cite{ben2019block} but no gains were obtained compared to feature concatenation. 

\begin{table}[]
\begin{center}
\small
\begin{tabular}{cc|c|c}
\multicolumn{1}{l}{\textbf{Projection}} & \multicolumn{1}{l|}{\textbf{Fusion}} & \multicolumn{1}{l|}{\textbf{Context}} & \multicolumn{1}{l}{\textbf{Bottles}} \\ \hline
Attention                               & MLB~\cite{kim2016hadamard}                                  & 80.83                                 & 78.26                                \\
Attention                               & Block~\cite{ben2019block}                                & 80.82                                 & 78.42                                \\
Attention                               & Concat                               & 81.09                                 & 78.45                                \\
GRU                                     & MLB~\cite{kim2016hadamard}                                  & 83.12                                 & 78.21                                \\
GRU                                     & Block~\cite{ben2019block}                                & 83.8                                  & 78.74                                \\
GRU                                     & Concat                               & 83.93                                 & 78.89                                \\
Avg Pooling                             & MLB~\cite{kim2016hadamard}                                  & 84.23                                 & 78.56                                \\
Avg Pooling                             & Block~\cite{ben2019block}                                & 85.11                                 & 79.15                                \\ \hline
\textbf{Avg Pooling}                    & \textbf{Concat}                      & \textbf{85.81}                        & \textbf{79.87}                      
\end{tabular}
\end{center}
\caption{Results obtained by employing different Projection and Fusion strategies on all the modules of our pipeline. Results are shown in terms of the mAP(\%). 
}
\label{tab:projection_fusion}
\vspace{-.5cm}
\end{table}

\subsection{Qualitative Results}
Qualitative results of the fine-grained image classification task are shown in Figure~\ref{fig:long}. By reviewing the samples obtained, we can note that our model is capable of learning a semantic space which combines successfully visual and textual signals coming from a single image. Classified samples such as "Pizzeria", "Tea House" and "Diner" often contain similar semantic classes ranked on second and third positions. Images belonging to the Drink Bottle dataset on the second row, are correctly classified even though text instances belong to specific brands, thus showing generalization capability of our method. The seventh image on the first row is wrongly classified as "Theatre" due to OCR recognition errors and a lack of strong enough visual cues. The remaining wrongly classified images are very challenging and contain some degree of ambiguity even for humans. 


\subsection{Fine-Grained Image Retrieval}
As an additional experiment that highlights the capabilities of the proposed model, we show the results obtained in Table~\ref{tab:retrieval_results} by performing query-by-example (QbE) image retrieval. In QbE, a system must return images in the form of a ranked list that belong to the same class as the image used as a query. To provide comparable results and following the work from~\cite{bai2018integrating, mafla2020fine}, we use the final classification vector as the image descriptor without using a specific metric-learning method. This vector is used to retrieve the nearest samples computed by the usage of the cosine similarity as a distance metric. 

In our experiments, the query, as well as the database is formed by unseen samples at training time.
\begin{table}[h]
\begin{center}
\small
\begin{tabular}{l|cc}
             \textbf{Method} & \textbf{Con-Text} & \textbf{Drink-Bottle} \\ \hline
Bai\cite{bai2018integrating}  & $62.87$ 

  & $60.80$                 \\
Mafla\cite{mafla2020fine}  & $64.52$                & $62.91$                 \\\hline
Proposed & \boldmath$75.50$            & \boldmath$65.39$                
\end{tabular}
\end{center}
\caption{Retrieval results on the evaluated datasets. The retrieval scores are depicted in terms of the mAP(\%).}
\label{tab:retrieval_results}
\end{table}
The results demonstrate that a very significant boost of $10.98$\% and $2.48$\% in Con-Text and Drink-Bottle is achieved respectively. 
The lower gain in the Drink-Bottle dataset directly depends on the harder to recognize text instances, as well as the low image quality of several samples that directly affects the model performance. 

Qualitative results that show the robustness of the model, as well as experiments addressing the importance of text can be found in the Supplementary Material section.

\section{Conclusions}
In this paper, we have presented a simple end-to-end model that employs a Multi-Modal Reasoning graph to encounter semantic and positional relationships between text and salient visual regions.
The learned space is composed by enriched features obtained from nodes in a graph, module that acts as an appropriate reasoning scheme.
Exhaustive experiments in two datasets and two different tasks validate the robustness of the presented model which achieve state-of-the-art results by a significant margin over previous methods. Moreover, our end-to-end pipeline does not require pre-computed handcrafted features or a collection of ensemble models as earlier works. 
In the future we expect to explore the effectiveness of this approach into other vision and language related tasks.


{\small
\bibliographystyle{ieee_fullname}
\bibliography{egbib}
}

\end{document}